%% file: main.tex
\documentclass[letterpaper]{article}

\usepackage{aaai25} 
\usepackage{times} 
\usepackage{helvet} 
\usepackage{courier} 
\usepackage[hyphens]{url} 
\usepackage{graphicx} 
\usepackage{graphicx}  
\usepackage{natbib}  
\usepackage{caption}  
\usepackage{algorithm}
\usepackage{algorithmic}

\usepackage{newfloat}
\usepackage{listings}
\DeclareCaptionStyle{ruled}{labelfont=normalfont,labelsep=colon,strut=off} 
\floatstyle{ruled}
\newfloat{listing}{tb}{lst}{}
\floatname{listing}{Listing}

\usepackage[table]{xcolor}
\definecolor{darkblue}{rgb}{0,0,0.5}

\usepackage{enumitem}
\usepackage{booktabs}
\usepackage{makecell}
\usepackage{multirow}
\usepackage{tabularray}

\usepackage{tikz}
\usetikzlibrary{positioning}
\usetikzlibrary{fit}

\begin{document}

\title{
    Large Language Models for Interpretable Mental Health Diagnosis
}
\author{
   Brian Hyeongseok Kim, Chao Wang \\
}
\affiliations {
   University of Southern California, Los Angeles, USA \\
   \texttt{\{brian.hs.kim, wang626\}@usc.edu}
}

\maketitle

\input{tex_new/0-abstract}
\input{tex_new/1-introduction}
\input{tex_new/2-preliminaries}

\input{tex_new/3-methodology}
\input{tex_new/4-evaluation}
\input{tex_new/5-related}
\input{tex_new/6-conclusion}

\bibliography{main}
\input{tex_new/appendix}

\end{document}

%% file: tex_new/0-abstract.tex
\begin{abstract}
We propose a clinical decision support system (CDSS) for mental health diagnosis that combines the strengths of large language models (LLMs) and constraint logic programming (CLP). 
Having a CDSS is important because of the high complexity of diagnostic manuals used by mental health professionals and the danger of diagnostic errors.
Our CDSS is a software tool that uses an LLM to translate diagnostic manuals to a logic program and solves the program using an off-the-shelf CLP engine to query a patient’s diagnosis based on the encoded rules and provided data.
By giving domain experts the opportunity to inspect the LLM-generated logic program, and making modifications when needed, our CDSS ensures that the diagnosis is not only accurate but also interpretable. 
We experimentally compare it with two baseline approaches of using LLMs: diagnosing patients using the LLM-only approach, and using the LLM-generated logic program but without expert inspection. 
The results show that, while LLMs are extremely useful in generating \emph{candidate} logic programs, these programs still require expert inspection and modification to guarantee faithfulness to the official diagnostic manuals.
Additionally, ethical concerns arise from the direct use of patient data in LLMs, underscoring the need for a safer hybrid approach like our proposed method.
\end{abstract}

%% file: tex_new/1-introduction.tex
\section{Introduction}

Mental disorders impose a significant burden on the affected individuals and their communities~\cite{burden_mh}.
Accurate diagnosis is a critical first step toward improving patient outcomes and fostering societal well-being.
In clinical settings, the diagnostic process relies on matching a patient's symptoms with the mental health diagnostic rules outlined in official manuals such as DSM-5-TR~\cite{dsm-5-tr} and ICD-11 CDDR~\cite{icd-11-cddr}.
These manuals, consisting of more than 1,000 pages of natural language descriptions, serve as authoritative references for not only mental health professionals but also insurance companies. 
However, their complexity poses a significant challenge, not only exacerbating the workload of already overburdened mental health professionals but also increasing the risk of diagnostic errors~\cite{apa_psych_burnout}.
This underscores the pressing need for developing a robust clinical decision support systems (CDSS), a software tool that can verify the diagnosis made manually.
Yet, such tools remain underdeveloped, particularly those that address issues of reliability and interpretability.

\input{tex_new/fig1}

Recent advancements in large language models (LLMs) suggest their potential in many applications~\cite{llm_applications} including clinical settings~\cite{llm_diagnosis_review}.
Thanks to their excellent processing and understanding of natural language, LLMs can generate diagnostic suggestions based on medical literature and patient data.
However, their adoption in clinical settings still faces challenges.
For example, LLMs are prone to issues like hallucinations~\cite{llm_hallucination, llm_hallucination_multimodal}, lack of explainability~\cite{llm_explainability} and consistency~\cite{llm_consistency}, and limited proficiency in complex reasoning~\cite{llm_reasoning}.
To date, no existing approach effectively combines LLMs with mechanisms that \emph{guarantee} accuracy and interpretability in the context of mental health diagnosis.

To fill this critical gap, we propose a method that combines LLMs with constraint logic programming (CLP), leading to a practical tool for assisting clinicians in making mental health diagnosis.
Specifically, our method leverages LLMs to translate natural language descriptions of mental health diagnostic criteria from manuals such as DSM-5-TR and ICD-11 CDDR to logic rules, thus reducing the cognitive burden on domain experts.
Simultaneously, we use an off-the-shelf CLP engine for solving the logic rules to ensure that the diagnostic output is verifiably correct, while enhancing interpretability through the rules and objectives explicitly defined via CLP.

Figure~\ref{fig:overview} shows the overall flow of our method. 
First, natural language text from a diagnostic manual (e.g., ICD-11 CDDR) is fed into an LLM, which generates a \emph{candidate} logic program codifying the diagnostic rules.
Next, a domain expert manually reviews the code to ensure that the LLM-generated rules accurately encode the manual's criteria.
Finally, the \emph{finalized} logic program is used by a CLP engine to generate the diagnosis result based on the information of an individual patient. 
By combining the natural language processing capabilities of an LLM with the logical reasoning capabilities of a CLP engine, our method delivers accurate and inherently interpretable diagnostic outcomes.

The rest of this paper is organized as follows: 
Section~\ref{sec:prelim} provides the background needed to contextualize our approach.
Section~\ref{sec:methodology} presents our methodology.
Section~\ref{sec:evaluation} presents the experimental results and analysis of our findings.
Section~\ref{sec:related} reviews the related work. Finally, Section~\ref{sec:conclusion} concludes with a summary of our contributions.

%% file: tex_new/fig1.tex
\begin{figure}[t]
\centering
\tikzstyle{arrow} = [thick,->,>=stealth]

\resizebox{\linewidth}{!}{
\begin{tikzpicture} [scale=1,text width=3.5cm, minimum height=1cm, align=center]

\tikzstyle{inputoutput} = [draw]
\tikzstyle{highlight} = [draw, fill=blue!10]
\tikzstyle{regular} = [draw, fill=gray!20]

\node[inputoutput] (manual)
{\textbf{Input}: ICD-11 CDDR Diagnostic Manual};

\node[highlight, below=1cm of manual] (llm)
{\textbf{LLM}:\\Generate Rules};

\node[regular, below=1cm of llm] (init_code)
{Candidate\\Logic Program};

\node[highlight, right=0.5cm of llm] (expert)
{\textbf{Expert}:\\Review Program};

\node[regular, below=1cm of expert] (final_code)
{Finalized\\Logic Program};

\node[highlight, right=0.5cm of expert] (lp)
{\textbf{CLP Engine}:\\Answer Query};

\node[inputoutput, above=1cm of lp] (input)
{\textbf{Input}: Individual Patient Information};

\node[inputoutput, below=1cm of lp] (output)
{\textbf{Output}:\\Diagnosis Result};

\node[draw, dashed, fit=(llm)(output), inner ysep=0.75cm, inner xsep=0.5cm] {};

\draw [arrow] (manual) -- (llm);
\draw [arrow] (llm) -- (init_code);
\draw [arrow] (init_code) -- (expert);
\draw [arrow] (expert) -- (final_code);
\draw [arrow] (final_code) -- (lp);
\draw [arrow] (input) -- (lp);
\draw [arrow] (lp) -- (output);

\end{tikzpicture}
}

\caption{Clinical decision support system (CDSS) combining the strengths of LLM and constraint logic programming.}
\label{fig:overview}
\end{figure}

%% file: tex_new/2-preliminaries.tex
\section{Background}
\label{sec:prelim}

In this section, we review the background information of psychological diagnosis and constraint logic programming.

\subsection{Psychological Diagnosis}

Psychological diagnosis is the process by which clinicians assess if a patient's symptoms meet the criteria for specific disorders as outlined in the diagnostic manuals such as DSM-5-TR  and ICD-11 CDDR.
%
%
%
These  authoritative manuals are widely adopted, underscoring their global relevance and importance.
As an example, consider the following ICD-11 CDDR diagnostic criteria for \emph{schizophrenia}:
\begin{quote}
{\footnotesize
\emph{At least two of the following symptoms must be present (by the individual’s report or through observation by the clinician or other informants) most of the time for a period of 1 month or more. 
At least one of the qualifying symptoms should be from items (a) to (d) below:
}

\emph{\textcolor{darkblue}{[List of symptoms from (a) to (g)... (omitted for brevity)]}.
}
}
\end{quote}

\noindent
Clinical decision support systems (CDSS) are a specific type of DSS~\cite{dss} where patient data and medical knowledge are integrated to the software tool, to assist clinicians with decision-making.
CDSS can address various scenarios such as offering diagnostic support, identifying drug interactions, and predicting treatment outcomes~\cite{cdss}.
The focus of this work is offering diagnostic support in the context of mental disorders.
Note that CDSS is merely a support system, as the name implies; it aims at helping clinical professionals in decision-making instead of replacing their decision-making role entirely.

\subsection{Constraint Logic Programming}

Constraint logic programming (CLP) is a paradigm that focuses on expressing logical rules of desired computations as opposed to implementing these computations.
%
%
It is well-suited for applications where accuracy and transparency are critical, as it focuses more on \emph{what} should be computed rather than \emph{how} to compute it, thus enabling easier verification of correctness and logical soundness.
In our work, we use Datalog as the CLP language, and solve Datalog programs using Soufflé, a state-of-the-art Datalog engine~\cite{souffle_cav16, souffle_cc16}.

\begin{listing}
\caption{An example logic program expressed in Datalog.}
\label{lst:edge_path}
\begin{lstlisting}
.decl Edge(x:number, y:number)
.decl Path(x:number, y:number)
.input Edge     
.output Path    
Path(x, y) :- Edge(x, y).
Path(x, y) :- Path(x, z), Edge(z, y).
\end{lstlisting}
\end{listing}

Listing~\ref{lst:edge_path} shows an example logic program that codifies the rules that infer the \texttt{Path} relation from the \texttt{Edge} relation. 
It starts with declarations of the two relations (Lines 1-2). Then, it specifies the input and the output (Lines 3-4). Finally, it defines the \emph{rules} for inferring \texttt{Path} from \texttt{Edge} (Lines 5-6). 
Specifically, Line~5 means that \texttt{Path(x,y)} holds if \texttt{Edge(x,y)} holds, and Line~6 means that \texttt{Path(x,y)} holds if both \texttt{Path(x,z)} and \texttt{Edge(z,y)} hold.
The comma (,) in Line 6 denotes logical AND, whereas a semicolon (;) denotes logical OR.

Given a set of \emph{facts}, e.g., \texttt{Edge} from 1 to 2 and \texttt{Edge} from 2 to 3, the program in Listing~\ref{lst:edge_path} computes all entries of the \texttt{Path} relation: from 1 to 2, from 2 to 3, and from 1 to 3.
This is how the program can answer queries, e.g., whether \texttt{Path(1,3)} holds. 
Similarly, we want to use Datalog to express ICD-11 CDDR diagnostic rules, and then answer queries for individual patients. 
This leads to a verifiably correct and explainable CDSS for mental disorders.

%% file: tex_new/3-methodology.tex
\section{Methodology}
\label{sec:methodology}

In this section, we present our Datalog encoding of diagnostic rules and LLM-based translation of text to rules. 

\subsection{Datalog Encoding of the Diagnosis}
\label{sec:methodology.datalog}

We focus on ICD-11 CDDR diagnostic rules, but this can be done similarly for DSM-5-TR.

\subsubsection{The Diagnostic Rules}

Listing~\ref{lst:diagnosis} shows a Datalog program with \emph{rules} that connect a patient's symptoms and past conditions to a mental disorder. Lines 4-6 specify the input and output relations.
For brevity, we omit the declarations of intermediate relations, but they also require the \texttt{.decl} keyword with variable types, similar to Lines 1-3.
The program first extracts the patient's name from \texttt{Observed} and add it to a relation called \texttt{AllPatients} (Line~7), and then identifies which symptoms are core (must be present) or qualifying (can be present) according to the diagnostic criteria.
Given a set of symptoms A, B, C, and D, for example, Symptoms A and B may be considered core whereas Symptoms C and D may be considered qualifying, and they must have been observed for more than 2 weeks (Lines~8-9).

\begin{listing*}
\caption{An example logic program for encoding ICD-11 CDDR diagnostic rules in Datalog.}
\label{lst:diagnosis}
\begin{lstlisting}
.decl Observed(Patient:symbol, Symptom:symbol, Week:float) 
.decl History(Patient:symbol, Condition:symbol, Count:number) 
.decl Diagnosis(Patient:symbol, Disorder:symbol) 
.input Observed
.input History
.output Diagnosis
AllPatients(P) :- Observed(P, _, _).    
Core(P, S, W) :- Observed(P, S, W), (S = "SymptomA"; S = "SymptomB"), Week>=2.
Qual(P, S, W) :- Observed(P, S, W), (S = "SymptomC"; S = "SymptomD"), Week>=2.    
CoreCount(P, count:Core(P, _, _)) :- Core(P, _, _).
CoreCount(P, 0) :-  !Core(P, _, _), AllPatients(P). 
QualCount(P, count:Qual(P, _, _)) :- Qual(P, _, _).
QualCount(P, 0) :-  !Qual(P, _, _), AllPatients(P). 
TotalCount(P, CC + QC) :- CoreCount(P, CC), QualCount(P, QC).   
Diagnosis(P, "DisorderD") :- CoreCount(P, CC), TotalCount(P, TC), History(P, "ConditionC", HC),  CC>=1, TC>=2, HC>=1.
\end{lstlisting}
\end{listing*}

The program has rules that count the number of symptoms in each category, which requires an aggregate function called \texttt{count} (Lines~10 and 12).
If \texttt{Core} or \texttt{Qual} relations do not exist, the count is set to 0 (Lines~11 and 13).
Once we have the counts for the symptoms, we add them up (Line~14).
Finally, we decide if a patient should be given the \texttt{Diagnosis} of ``\texttt{DisorderD}"
 (Line~15).
Here, the diagnosis requires at least 1 core symptom and at least 2 symptoms in total (i.e., one core and one qualifying, or two core symptoms), and at least one occurrence of ``\texttt{ConditionC}" in prior history.

\subsubsection{Patient Information}

The Datalog program in Listing~\ref{lst:diagnosis} requires  \emph{facts} that describe the patient as input.
These facts are expressed using the \texttt{Observed} and \texttt{History} relations. 
\texttt{Observed} indicates that \texttt{Patient} is experiencing \texttt{Symptom} for the duration of \texttt{Week} (Line 1).
\texttt{History} indicates that \texttt{Patient} has a history of \texttt{Condition} for the \texttt{Count}\footnote{Note that the lowercase \texttt{count} refers to the aggregate function, while the uppercase \texttt{Count} refers to the variable name.} number of times (Line 2).

Consider the follwing example of ``PatientA'', who has been observed with ``SymptomA'' and ``SymptomB'' for 3.5 weeks, and has a prior history of ``ConditionC'' two times.
The corresponding input facts are as follows: 
\begin{itemize}
\item \texttt{Observed}(``PatientA", ``SymptomA", 3.5)
\item \texttt{Observed}(``PatientA", ``SymptomB", 3.5)
\item \texttt{History}(``PatientA", ``ConditionC", 2)
\end{itemize}%
They meet the diagnostic criteria for ``{DisorderD}" as shown in Line 15 of Listing~\ref{lst:diagnosis}.
%
%

\subsection{LLM-Based Translation of Manuals to Rules}
\label{sec:methodology.llm}

We prompt LLMs to translate the text-based diagnostic criteria from the ICD-11 CDDR manual into \emph{candidate} logic programs in Datalog, similar to the program shown in Listing~\ref{lst:diagnosis}.
Then, we assess whether the LLM-generated logic program can diagnose a given patient correctly.
In-context learning (ICL)~\cite{llm_icl, llm_icl_survey} allows LLMs to perform tasks better without explicitly updating the model parameters.
As part of ICL, we provide an example of diagnostic criteria text from ICD-11 CDDR and its corresponding Datalog program, such that the models can learn from the demonstrated task.
Our one-shot prompt template is as follows:
\begin{quote}
{\footnotesize
\emph{
\textbf{System}: 
You are an expert at translating mental health diagnostic criteria into a Datalog program in Soufflé.
The patient data is given as input to the program as \texttt{Observed} and \texttt{History} relations. 
The patient diagnosis is returned as output from the program as \texttt{Diagnosis} relation.
\textcolor{darkblue}{Explain the relations.}
}

\emph{
\textbf{Example}: 
\textcolor{darkblue}{Include the ICD-11 CDDR diagnostic criteria for a disorder and its corresponding Datalog program.}
}

\emph{
\textbf{Task}: 
Translate the given criteria into a Datalog program using Soufflé syntax. 
\textcolor{darkblue}{Include relevant \texttt{Observed} symptom names, \texttt{History} condition names, and the ICD-11 CDDR diagnostic criteria for each disorder.}
}
}
\end{quote}

\noindent
Since the generated programs are declarative, and they are driven by logic, the diagnoses that they provide are guaranteed to be correct, as long as the rules reflect the logic of the diagnostic manual accurately.
While LLMs may produce \emph{candidate} logic programs that contain syntactic and/or semantic errors, these logic programs may be reviewed and corrected by a domain expert.
At the level of Datalog programs, expert intervention is feasible and sufficient for ensuring that the \emph{finalized} logic programs not only can be compiled, but also accurately represent the diagnostic criteria.
Furthermore, manual inspection is reasonable in this context, given the critical role of human oversight in clinical applications.
We refer the readers to Appendix~\ref{sec:appendix_prompts} for the detailed prompts used in our experimental evaluation.

%% file: tex_new/4-evaluation.tex
\section{Evaluation}
\label{sec:evaluation}

We use ICD-11 CDDR diagnostic manual~\cite{icd-11-cddr} and focus on four mood disorders: Bipolar I (BPD1), Bipolar II (BPD2), Single Episode Depressive Disorder (SEDD), and Recurrent Depressive Disorder (RDD).
From natural language descriptions of the diagnostic criteria, our method generates the \emph{candidate} Datalog program as described in Section~\ref{sec:methodology.datalog}.  
Then, we manually inspect the LLM-generated Datalog program and correct errors to ensure that the \emph{finalized} Datalog program accurately encodes the diagnostic rules.

We also manually validated the diagnosis results of the Datalog program. Given a dataset of 30 patients, the finalized Datalog program identified 9 patients with BPD1, 8 with BPD2, 5 with SEDD, and 4 with RDD.
Four patients remained undiagnosed, as they did not meet the criteria for any of the considered mood disorders.
We validated the program's results by cross-checking the patient data against the diagnostic criteria specified in the manual.
Details of all 30 patients can be found in Appendix~\ref{sec:appendix_patients}.

\subsection{The Experimental Setup}

Given the time and effort required for manual rule translation and diagnosis validation, we aim to assess whether state-of-the-art LLMs can automate this process without compromising accuracy.
Our main research questions are:
\begin{enumerate}[label=RQ\arabic*., leftmargin=*]
\item How accurate are the diagnostic outputs generated by the LLM-translated programs?
\item To what extent can LLMs accurately interpret and translate diagnostic criteria from text into Datalog?
\item How much additional human effort is required to correct errors in the LLM-translated programs?
\item How effective are LLMs in diagnosing a patient when given their data directly?
\end{enumerate}

\noindent
We used 3 state-of-the-art LLMs.
%
%
\textbf{GPT} stands for \textsc{GPT-4o} on OpenAI, released in May 2024~\cite{gpt4_openai_2024}.\footnote{\url{https://chatgpt.com}}
\textbf{Gemini} stands for \textsc{Gemini-1.5-Flash} on Google Cloud, released in May 2024~\cite{gemini1.5_google_2024}\footnote{\url{https://gemini.google.com}}.
\textbf{Llama} stands for \textsc{Llama-3.2} on Meta AI, released in September 2024~\cite{llama3_meta_2024}.\footnote{\url{https://www.meta.ai}}
%
These LLMs were accessed between October and November 2024.

\subsection{Two Baseline Approaches}

To evaluate our method, we developed two groups of baselines for comparison.
The first group of baselines (\emph{LLM-only}) involves directly providing a diagnosis given the patient data.
This is analogous to using LLMs as an external consultant that can either validate or challenge a clinician's diagnosis~\cite{llm_diagnosis_consultant}.
For this task, we use the following prompt template:
\begin{quote}
{\footnotesize
\emph{
\textbf{System}: 
You are an expert at diagnosing patients according to the ICD-11 CDDR.
The considered disorders are \textcolor{darkblue}{[List of mood disorders]}.
The patient data is given as \texttt{Observed} and \texttt{History} relations.
\textcolor{darkblue}{Explain the relations.}
}

\emph{
\textbf{Task}: 
Please output the diagnosis for the following patients.
Patients with no clear diagnosis should be indicated as such.
\textcolor{darkblue}{Include patient data.}
}
}
\end{quote}

\noindent
The second group of baselines (\emph{LLM + Datalog}) is running the LLM-translated Datalog programs without expert intervention.
This approach is comparable to testing LLM-generated code in imperative programming~\cite{llm_codegen}.
These baselines follow the same prompt structure described in Section~\ref{sec:methodology.llm}.
Our method extends these baselines by incorporating expert corrections to address syntactic and semantic errors in the LLM-generated code.
We refer to these expert-corrected programs as \emph{Our CDSS}.


\subsection{Experimental Results}

\begin{table*}[t]
\centering
\caption{
Comparing our method with baselines on the first 10 (out of 30) patients. 
`Known Disorder' indicates what the patient is diagnosed with according to the ICD-11 CDDR criteria. 
`LLM-only' indicates the diagnosis directly produced by LLMs.
`LLM + Datalog' indicates the diagnosis produced by the LLM-generated Datalog program.
`Our CDSS' indicates the diagnosis produced by our method. 
The symbol `-' indicates no clear diagnosis.
}
\label{tab:main}

\footnotesize
\begin{tabular}{cc ccc ccc c}
\toprule
\multirow{2}{*}{\textbf{Patient ID}}
& \multirow{2}{*}{\textbf{Known Disorder}}
& \multicolumn{3}{c}{\emph{Diagnosis by LLM-only Approach} }
& \multicolumn{3}{c}{\emph{Diagnosis using LLM + Datalog}}
& \emph{Diagnosis by Our CDSS}
\\
\cmidrule(lr){3-5} \cmidrule(lr){6-8} \cmidrule{9-9}
& &
\textbf{Llama}
& \textbf{Gemini}
& \textbf{GPT}
& \textbf{Llama}
& \textbf{Gemini}
& \textbf{GPT}
& \textbf{GPT}
\\
\midrule
{No. 1} & BPD2      
        & \cellcolor{lime}BPD2
        & BPD1
        & \cellcolor{lime}BPD2
        & -
        & -
        & \cellcolor{lime}BPD2
        & \cellcolor{lime}BPD2
        \\
{No. 2} & RDD  
        & SEDD
        & SEDD
        & SEDD
        & BPD1
        & SEDD
        & SEDD
        & \cellcolor{lime}RDD
        \\
{No. 3} & BPD1      
        & \cellcolor{lime}BPD1
        & \cellcolor{lime}BPD1
        & \cellcolor{lime}BPD1
        & \cellcolor{lime}BPD1
        & \cellcolor{yellow} BPD1, BPD2
        & \cellcolor{lime}BPD1
        & \cellcolor{lime}BPD1
        \\
{No. 4} & BPD2
        & SEDD
        & \cellcolor{lime}BPD2
        & \cellcolor{lime}BPD2
        & BPD1
        & SEDD
        & -
        & \cellcolor{lime}BPD2
        \\
{No. 5} & BPD1     
        & \cellcolor{lime}BPD1
        & \cellcolor{lime}BPD1
        & \cellcolor{lime}BPD1
        & \cellcolor{lime}BPD1
        & \cellcolor{yellow} BPD1, BPD2
        & -
        & \cellcolor{lime}BPD1
        \\
{No. 6} & BPD2      
        & \cellcolor{lime}BPD2
        & \cellcolor{lime}BPD2
        & \cellcolor{lime}BPD2
        & BPD1
        & SEDD
        & \cellcolor{lime}BPD2
        & \cellcolor{lime}BPD2
        \\
{No. 7} & BPD1      
        & -
        & \cellcolor{lime}BPD1
        & \cellcolor{lime}BPD1
        & -
        & \cellcolor{lime}BPD1
        & \cellcolor{lime}BPD1
        & \cellcolor{lime}BPD1
        \\
{No. 8} & SEDD
        & \cellcolor{lime}SEDD
        & \cellcolor{lime}SEDD
        & \cellcolor{lime}SEDD
        & BPD1
        & -
        & \cellcolor{lime}SEDD
        & \cellcolor{lime}SEDD
        \\
{No. 9} & SEDD      
        & \cellcolor{lime}SEDD
        & \cellcolor{lime}SEDD
        & \cellcolor{lime}SEDD
        & BPD1
        & -
        & \cellcolor{lime}SEDD
        & \cellcolor{lime}SEDD
        \\
{No. 10} & -        
        & \cellcolor{lime}-
        & \cellcolor{lime}-
        & \cellcolor{lime}-
        & \cellcolor{lime}-
        & \cellcolor{lime}-
        & \cellcolor{lime}-
        & \cellcolor{lime}-
        \\
\midrule
\multicolumn{2}{c}{\textbf{Correct Diagnosis (Total):}} 
& 7/10
& 8/10
& 9/10
& 3/10
& (2+2)/10
& 7/10
& 10/10
\\
\bottomrule
\end{tabular}

\end{table*}

%
Table~\ref{tab:main} compares the performance of our method against the baselines across 10 patients.
The remainder of this section will discuss the results for the first 10 patients.
Results for all 30 patients can be found in Appendix~\ref{sec:appendix_results}.

Columns~1-2 list patient numbers and their disorders based on our manually written Datalog program, validated against the ICD-11 CDDR criteria.
Columns~3-5 (\emph{LLM-only}) show diagnoses directly provided by the LLMs, while Columns~6-8 (\emph{LLM + Datalog}) show diagnoses from LLM-generated Datalog programs without expert intervention.
Column~9 (\emph{Our CDSS}) shows diagnoses from an expert-corrected LLM-generated program.
Green cells indicate correct diagnoses, yellow cells indicate partial correctness (correct diagnosis with additional incorrect ones), and the final row summarizes correct diagnoses per method.

\paragraph{Answer to RQ1}
To address RQ1 on the accuracy of LLM-translated programs, we look at \emph{LLM + Datalog} columns for the as-is versions\footnote{
The GPT-generated program compiled and ran as-is, while programs from Gemini and Llama required minor syntactic fixes for compilation.
For Gemini, 17 lines were added, and 4 removed from an initial 34 lines, resuling in 47 lines of code (LoC).
For Llama, 19 lines were added, and 11 removed from an initial 114 lines, resulting in 122 LoC.
No semantic changes were made.
}
%
%
%
%
%
%
and the \emph{Our CDSS} column for the expert-corrected version.
Among the as-is programs, GPT performs the best with 7 correct diagnoses out of 10, followed by Gemini with 2 correct and 2 partially correct, and Llama with 3 correct.
This pattern holds across all the patients, where GPT achieves 22 correct diagnoses out of 30, while Gemini has 8 correct and 4 partially correct, and Llama only 9 correct.
We extend the most accurate program generated by GPT and implement logical changes to align with the ICD-11 CDDR criteria for the mood disorders.
The expert-reviewed program, shown in Column 9, produces 10 correct diagnoses out of 10 (30 out of 30).

\paragraph{Answer to RQ2}
To address RQ2 on the performance of LLMs in translating diagnostic criteria into Datalog programs, we take a closer look at the programs that correspond to Columns~6-8.

Although GPT-generated program achieves the greatest number of correct diagnoses, it relies solely on \texttt{History} in its final diagnostic rules, despite constructing intermediate rules to identify mood episodes based on \texttt{Observed}.
This issue arises from the diagnostic text's phrasing, which specifies a ``history of" certain mood episodes as a requirement for a particular mood disorder.
While clinicians would intuitively consider current symptoms for diagnosis, GPT interprets the text literally, lacking the nuanced understanding needed for accurate clinical interpretation.

The Gemini-generated code presents the opposite issue, where it only considers current symptoms, ignoring the patient's history.
This leads to missed diagnoses such as Patient 1, where the current symptoms suggest no diagnosis (denoted `-' under Column~7), despite the patient's history indicating BPD2.
This may stem from using a Datalog program for schizophrenia in the prompt, which doesn't incorporate \texttt{History}, which is specific to mood disorders.
Additionally, Gemini-generated program frequently diagnoses conflicting disorders (e.g., ``BPD1, BPD2" for Patients 3 and 5), which contradicts the diagnostic criteria that require mutually exclusive conditions.
BPD1 requires a mixed or manic episode; BPD2 explicitly requires the absence of such episodes.

The Llama-generated code often misdiagnoses patients as BPD1 (e.g., Patients 2, 4, 6, 8, and 9).
This stems from a lack of intermediary logic to properly identify mood episodes.
The generated program counts associated symptoms without distinguishing core or qualifying symptoms and uses an arbitrary threshold that doesn't align with the ICD-11 criteria, leading to frequent BPD1 diagnoses, as it has the lowest threshold.

Overall, the inconsistency across models suggests that while LLM-generated code shows promise, it is not yet reliable for direct clinical use.
Models could benefit from more text-to-rule translation examples for ICL to generate more accurate programs.
Additionally, breaking down the task into smaller steps through multi-turn conversation~\cite{llm_multi_turn} or Chain-of-Thought (CoT) prompting~\cite{llm_cot} could enhance their logical reasoning.
Models could be fine-tuned for this task by experts with reinforcement learning from human feedback (RLHF)~\cite{llm_rlhf}.
Finally, using LLMs optimized for code generation (e.g., GitHub Copilot\footnote{https://copilot.github.com/} and Amazon Q Developer\footnote{https://aws.amazon.com/q/developer/}) could improve the performance.

\paragraph{Answer to RQ3.}

\begin{listing*}
\caption{
Portion of manually corrected Datalog code in the candidate logic program generated by LLM.
}
\label{lst:code_evolution}
\begin{lstlisting}
DepressiveEpisode(Patient) :- &\newline& &\textcolor{green!40!black}{+ !\textbf{MixedEpisode}(Patient),}& &\newline& DepressiveSymptomCount(Patient, Count), Count >= 5, AffectiveCluster(Patient, _).  
MixedEpisode(Patient) :- &\newline& &\textcolor{red}{- \textbf{ManicEpisode}(Patient), \textbf{DepressiveEpisode}(Patient).}& &\newline& &\textcolor{green!40!black}{+ \textbf{DepressiveSymptomCount}(Patient, DepressiveCount), DepressiveCount >= 3, \textbf{MixedManicSymptomCount}(Patient, ManicCount), ManicCount >= 3, \textbf{MixedCore}(Patient).}&    
Diagnosis(Patient, "Bipolar_I") :- History(Patient, "manic_episode", Count1), Count1 >= 1.
Diagnosis(Patient, "Bipolar_I") :- History(Patient, "mixed_episode", Count2), Count2 >= 1.
&\textcolor{green!40!black}{+ \textbf{Diagnosis}(Patient, "Bipolar\_I") :- \textbf{ManicEpisode}(Patient); \textbf{MixedEpisode}(Patient).}&
\end{lstlisting}
\end{listing*}

As discussed, LLM-generated programs do not guarantee that the encoded logic accurately replicates the diagnostic criteria of ICD-11 CDDR.
%
%
To address this, our method proposes a pipeline where LLMs generate candidate Datalog programs, which are then reviewed and refined by experts.
This approach aims to balance the efficiency of AI with the crucial need for diagnostic accuracy.
In this context, it is important to address RQ3, which examines the additional human effort required to accurately represent the logic of the diagnostic criteria.

Listing~\ref{lst:code_evolution} highlights some of the changes made to the GPT-generated code, addressing two major issues.
First, the original definition of \texttt{MixedEpisode} (Line 2) created a cyclic dependency by requiring both \texttt{ManicEpisode} and \texttt{DepressiveEpisode} for its definition.
However, according to the ICD-11 CDDR manual, a mixed episode should be defined independently of these episodes and based on specific symptom thresholds.
Furthermore, the logic expressed in the program contradicted clinical guidelines, since depressive or manic episodes should not apply if the symptoms qualify better as a mixed episode.
The revised code resolves this by incorporating the absence of a mixed episode as part of criteria for other mood episodes (Line~1) and redefining \texttt{MixedEpisode} to directly evaluate symptom counts and core criteria (Line~2).
This approach eliminates the cyclic dependency and ensures compliance with clinical guidelines.
In order to achieve this, we manually added several missing intermediate relations and rules to accurately identify and count core and qualifying symptoms for different mood episodes.

Second, the original logic for diagnosing disorders relied solely on \texttt{History} (Lines~3-4).
As discussed, this approach neglected the possibility of diagnosing based on present mood episodes.
The corrected code addresses this by checking for the presence of the current mood episodes based on \texttt{Observed} symptoms (Line~5).
Overall, these corrections ensure that the diagnosis logic is more comprehensive and aligns with clinical practice.

The final corrected version of the GPT-generated code passes for all 10 (30) patients.
In total, 57 lines were added and 10 removed from the initial 107 lines of code (LoC), resulting in a final 154 LoC.
The first set of corrections—addressing cyclic dependencies and clinical inconsistencies—required the addition of 47 LoC and removal of 6, reflecting changes that demanded significant domain expertise.
In contrast, modifying the diagnosis logic to incorporate present mood episodes added 10 LoC and removed 4, which were relatively straightforward adjustments.

These statistics highlight the varying levels of effort needed to refine different aspects of the generated code, while also demonstrating that much of the initial code was functional and required only minimal deletions to align with clinical guidelines.
Despite the manual effort required, LLMs significantly accelerate the initial code generation process, providing a strong foundation that would otherwise require substantial time and expertise to build from scratch.

\paragraph{Answer to RQ4.}
To answer RQ4, which evaluates the effectiveness of LLMs in diagnosing patients directly, we revisit Table~\ref{tab:main} under the \emph{LLM-only} columns.
Among the tested models, GPT leads with 9 correct diagnoses out of 10, followed by Gemini with 8 and Llama with 7.
This trend extends across all patients, where GPT leads with 22 correct diagnoses out of 30, followed by Gemini and Llama with 19 each.

Directly using LLMs for diagnosis generally results in higher accuracy than relying on LLM-generated candidate programs.
However, the variability in model performance highlights significant challenges.
LLMs inherently rely on probabilistic predictions rather than logical proofs, making it difficult to guarantee consistency and accuracy required in medical contexts.
Furthermore, their complex architectures make them hard to interpret.
Unlike LLM-generated logic programs, which offer transparent reasoning steps, the direct diagnoses provided by LLMs remain opaque, even when correct.
Finally, there are always ethical implications and privacy concerns of providing real patient data to LLMs, which complicates their direct application in healthcare.

Instead, our proposed method of combining LLMs with constraint logic programming offers a promising alternative.
LLMs can be leveraged to generate interpretable logical rules that determine if a patient meets specific diagnostic criteria.
This approach reduces ethical concerns by avoiding direct input of sensitive patient data into LLMs and takes advantage of the increasing availability of code-generative LLMs.
Moreover, logic programs are inherently transparent and interpretable, as their rules can be manually verified for alignment with diagnostic standards of the manuals. 
%
%

%% file: tex_new/5-related.tex
\section{Related Work}
\label{sec:related}

To the best of our knowledge, our proposed method is the first method for combining LLMs and constraint logic programming to provide a clinical decision support system (CDSS) in the context of mental health diagnosis.
Other recent studies have explored the use of LLMs in mental health contexts, including developing chat-based counselors~\cite{llm_mh_chatcounselor}, analyzing emotions~\cite{llm_mh_emotion_analysis}, and predicting mental states from online text~\cite{llm_mh_predict_online}.
However, these works do not extend to creating diagnostic tools or integrating logic-based reasoning to support clinical decision-making.

Beyond clinical applications, there is ongoing research on LLMs for logical reasoning, such as translating text into specifications for Boolean satisfiability (SAT) solvers~\cite{llm_isil_sat} or evaluating their reasoning capabilities in mathematical and strategic domains~\cite{llm_math, llm_strategic}.
While these efforts demonstrate LLMs’ potential for logic-based tasks, they do not address the use of logic programming languages like Datalog in clinical settings.

Prior work on using logic in CDSS take an ontological approach of structured knowledge representations for diagnosis~\cite{psydis_2012}, or apply satisfiability modulo theory (SMT) solvers and theorem provers to detect conflicts in medical treatments~\cite{bowles_2019}.
While these works address relevant clinical needs, they precede the advent of modern LLMs and do not incorporate logic programming. 
Our work bridges these gaps by leveraging LLMs to generate interpretable logic programs for mental health diagnosis, offering a unique combination of efficient AI techniques and explainable logic-based computation to assist clinicians.

%% file: tex_new/6-conclusion.tex
\section{Conclusion}
\label{sec:conclusion}

We present a novel approach that integrates large language models (LLMs) with constraint logic programming (CLP) to design a clinical decision support systems (CDSS) for mental health diagnosis.
Our evaluation demonstrates that while LLMs show promise for diagnostic tasks, they still face significant limitations when used directly, including issues with consistency, interpretability, and ethical concerns related to patient data.
To address these challenges, our method utilizes LLMs to generate logic programs that encode diagnostic rules, and CLP engines to produce diagnostic results based on patient data.
We propose that this hybrid approach, combined with expert validation, ensures that diagnostic reasoning is aligned with clinical criteria, enhancing reliability and safety in clinical decision-making.
%
%
Future work will explore domain-specific fine-tuning of LLMs, evaluate the approach on real-world datasets, and extend the Datalog encoding to address more nuanced diagnostic criteria and specifiers.

\section*{Ethical Statement}
The proposed CDSS aims at helping clinical professionals in decision-making.  It is not meant to replace or refute the diagnoses provided by qualified clinicians. 
All evaluations and decisions regarding diagnoses must be conducted in accordance with the expertise of trained professionals. 
The hypothetical data and modeling used in this study are intended for proof of concept and are not meant to substitute for real patient data, which may be more complex.

%% file: tex_new/appendix.tex
\clearpage\newpage
\onecolumn

\appendix

\section{Detailed Diagnosis Results}
\label{sec:appendix_results}

\begin{table*}[ht]
\centering
\caption{Comparing our method with baselines on all 30 patients. 
`Known Disorder' indicates what the patient is diagnosed with according to the ICD-11 CDDR criteria. 
`LLM-only' indicates the diagnosis directly produced by LLMs. 
`LLM+Datalog' indicates the diagnosis produced by the LLM-generated Datalog program. 
`Our CDSS' indicates the diagnosis produced by our method. 
The symbol `-' indicates no clear diagnosis. 
}

{\footnotesize
\begin{tabular}{cc ccc ccc ccc}
\toprule
\multirow{2}{*}{\textbf{Patient ID}}
& \multirow{2}{*}{\textbf{Known Disorder}}
& \multicolumn{3}{c}{\emph{Diagnosis by LLM-only Approach}}
& \multicolumn{3}{c}{\emph{Diagnosis using LLM + Datalog}}
& \emph{Diagnosis by Our CDSS}
\\
\cmidrule(lr){3-5} \cmidrule(lr){6-8} \cmidrule{9-9}
&
& \textbf{Llama}
& \textbf{Gemini}
& \textbf{GPT}
& \textbf{Llama}
& \textbf{Gemini}
& \textbf{GPT}
& \textbf{GPT}
\\
\midrule
{No. 1} & BPD2    
        & \cellcolor{lime}BPD2
        & BPD1
        & \cellcolor{lime}BPD2
        & -
        & -
        & \cellcolor{lime}BPD2
        & \cellcolor{lime}BPD2
        \\
{No. 2} & RDD
        & SEDD
        & SEDD
        & SEDD
        & BPD1
        & SEDD
        & SEDD
        & \cellcolor{lime}RDD
        \\
{No. 3} & BPD1
        & \cellcolor{lime}BPD1
        & \cellcolor{lime}BPD1
        & \cellcolor{lime}BPD1
        & \cellcolor{lime}BPD1
        & \cellcolor{yellow} BPD1, BPD2
        & \cellcolor{lime}BPD1
        & \cellcolor{lime}BPD1
        \\
{No. 4} & BPD2
        & SEDD
        & \cellcolor{lime}BPD2
        & \cellcolor{lime}BPD2
        & BPD1
        & SEDD
        & -
        & \cellcolor{lime}BPD2
        \\
{No. 5} & BPD1
        & \cellcolor{lime}BPD1
        & \cellcolor{lime}BPD1
        & \cellcolor{lime}BPD1
        & \cellcolor{lime}BPD1
        & \cellcolor{yellow} BPD1, BPD2
        & -
        & \cellcolor{lime}BPD1
        \\
{No. 6} & BPD2
        & \cellcolor{lime}BPD2
        & \cellcolor{lime}BPD2
        & \cellcolor{lime}BPD2
        & BPD1
        & SEDD
        & \cellcolor{lime}BPD2
        & \cellcolor{lime}BPD2
        \\
{No. 7} & BPD1
        & -
        & \cellcolor{lime}BPD1
        & \cellcolor{lime}BPD1
        & -
        & \cellcolor{lime}BPD1
        & \cellcolor{lime}BPD1
        & \cellcolor{lime}BPD1
        \\
{No. 8} & SEDD
        & \cellcolor{lime}SEDD
        & \cellcolor{lime}SEDD
        & \cellcolor{lime}SEDD
        & BPD1
        & -
        & \cellcolor{lime}SEDD
        & \cellcolor{lime}SEDD
        \\
{No. 9} & SEDD
        & \cellcolor{lime}SEDD
        & \cellcolor{lime}SEDD
        & \cellcolor{lime}SEDD
        & BPD1
        & -
        & \cellcolor{lime}SEDD
        & \cellcolor{lime}SEDD
        \\
{No. 10} & -
        & \cellcolor{lime}-
        & \cellcolor{lime}-
        & \cellcolor{lime}-
        & \cellcolor{lime}-
        & \cellcolor{lime}-
        & \cellcolor{lime}-
        & \cellcolor{lime}-
        \\
\hline
{No. 11} & -
        & BPD2
        & BPD2
        & BPD2
        & BPD1
        & \cellcolor{lime}-
        & \cellcolor{lime}-
        & \cellcolor{lime}-
        \\
{No. 12} & BPD1
        & \cellcolor{lime}BPD1
        & \cellcolor{lime}BPD1
        & BPD2
        & \cellcolor{lime}BPD1
        & \cellcolor{lime}BPD1
        & -
        & \cellcolor{lime}BPD1
        \\
{No. 13} & BPD1
        & \cellcolor{lime}BPD1
        & \cellcolor{lime}BPD1
        & \cellcolor{lime}BPD1
        & \cellcolor{lime}BPD1
        & \cellcolor{lime}BPD1
        & \cellcolor{lime}BPD1
        & \cellcolor{lime}BPD1
        \\
{No. 14} & BPD1
        & \cellcolor{lime}BPD1
        & \cellcolor{lime}BPD1
        & \cellcolor{lime}BPD1
        & \cellcolor{lime}BPD1
        & \cellcolor{yellow} BPD1, BPD2
        & \cellcolor{lime}BPD1
        & \cellcolor{lime}BPD1
        \\
{No. 15} & BPD2
        & \cellcolor{lime}BPD2
        & \cellcolor{lime}BPD2
        & \cellcolor{lime}BPD2
        & BPD1
        & BPD1
        & RDD
        & \cellcolor{lime}BPD2
        \\
{No. 16} & BPD2
        & \cellcolor{lime}BPD2
        & \cellcolor{lime}BPD2
        & \cellcolor{lime}BPD2
        & BPD1
        & BPD1
        & \cellcolor{lime}BPD2
        & \cellcolor{lime}BPD2
        \\
{No. 17} & BPD1
        & \cellcolor{lime}BPD1
        & \cellcolor{lime}BPD1
        & \cellcolor{lime}BPD1
        & \cellcolor{lime}BPD1
        & \cellcolor{lime}BPD1
        & \cellcolor{lime}BPD1
        & \cellcolor{lime}BPD1
        \\
{No. 18} & RDD
        & \cellcolor{lime}RDD
        & SEDD
        & \cellcolor{lime}RDD
        & BPD1
        & -
        & \cellcolor{lime}RDD
        & \cellcolor{lime}RDD
        \\
{No. 19} & BPD2      
        & \cellcolor{lime}BPD2
        & \cellcolor{lime}BPD2
        & \cellcolor{lime}BPD2
        & BPD1
        & SEDD
        & \cellcolor{lime}BPD2
        & \cellcolor{lime}BPD2
        \\
{No. 20} & SEDD
        & -
        & -
        & -
        & -
        & -
        & \cellcolor{lime}SEDD
        & \cellcolor{lime}SEDD
        \\
{No. 21} & BPD1
        & \cellcolor{lime}BPD1
        & \cellcolor{lime}BPD1
        & \cellcolor{lime}BPD1
        & \cellcolor{lime}BPD1
        & -
        & \cellcolor{lime}BPD1
        & \cellcolor{lime}BPD1
        \\
{No. 22} & BPD1
        & \cellcolor{lime}BPD1
        & \cellcolor{lime}BPD1
        & \cellcolor{lime}BPD1
        & \cellcolor{lime}BPD1
        & \cellcolor{lime}BPD1
        & \cellcolor{lime}BPD1
        & \cellcolor{lime}BPD1
        \\
{No. 23} & BPD2     
        & -
        & BPD1
        & -
        & -
        & BPD1
        & SEDD
        & \cellcolor{lime}BPD2
        \\
{No. 24} & -
        & BPD1
        & BPD1
        & BPD1
        & BPD1
        & BPD1
        & \cellcolor{lime}-
        & \cellcolor{lime}-
        \\
{No. 25} & RDD
        & \cellcolor{lime}RDD
        & SEDD
        & \cellcolor{lime}RDD
        & BPD1
        & -
        & \cellcolor{lime}RDD
        & \cellcolor{lime}RDD
        \\
{No. 26} & BPD2
        & SEDD
        & BPD1
        & \cellcolor{lime}BPD2
        & BPD1
        & \cellcolor{yellow} BPD1, BPD2
        & \cellcolor{lime}BPD2
        & \cellcolor{lime}BPD2
        \\
{No. 27} & SEDD
        & RDD
        & \cellcolor{lime}SEDD
        & RDD
        & BPD1
        & -
        & \cellcolor{lime}SEDD
        & \cellcolor{lime}SEDD
        \\
{No. 28} & SEDD
        & \cellcolor{lime}SEDD
        & \cellcolor{lime}SEDD
        & \cellcolor{lime}SEDD
        & BPD1
        & \cellcolor{lime}SEDD
        & -
        & \cellcolor{lime}SEDD
        \\
{No. 29} & RDD     
        & SEDD
        & SEDD
        & SEDD
        & BPD1
        & SEDD
        & SEDD
        & \cellcolor{lime}RDD
        \\
{No. 30} & -        
        & BPD2
        & BPD2
        & \cellcolor{lime}-
        & BPD1
        & BPD1
        & \cellcolor{lime}-
        & \cellcolor{lime}-
        \\
\midrule
\multicolumn{2}{c}{\textbf{Correct Diagnosis (Total):}} 
& 19/30
& 19/30
& 22/30
& 9/30
& (8+4)/30
& 22/30
& 30/30
\\
\bottomrule
\end{tabular}
}
\end{table*}

\newpage
\section{Patient Information}
\label{sec:appendix_patients}

\begin{longtblr}[
caption = {
Input data of all 30 patients. 
The symbol `-' for Column 4 indicates that there is no prior history condition of mood episode.
`-' in Column 5 indicates that the patients' observed symptoms do not qualify for a current mood episode.
},
]{
colspec = {cclll},
hline{1,Z} = {0.05em},
hline{2} = {0.03em},
row{1} = {font=\bfseries},
rowhead = 1,
}
{Patient ID} & {Disorder} & {Observed Symptoms} & {History Conditions} & {Mood Episode} \\

{No. 1} & BPD2
        & \makecell[l]{
        depressed\_mood 1.5 \\ reduced\_concentration 1.2 \\ reduced\_energy 0.8 \\ increased\_talkativeness 0.6
        }
        & \makecell[l]{depressive 1 \\ hypomanic 1}
        & -
        \\
        
{No. 2} & RDD
        & \makecell[l]{
        depressed\_mood 5.7 \\ 
        diminished\_interest\_pleasure 5.7 \\ 
        reduced\_concentration 3.5 \\ 
        low\_self\_worth 2.0 \\ 
        psychomotor\_disturbances 5.7
        }
        & depressive 1
        & depressive
        \\

{No. 3} & BPD1
        & \makecell[l]{
        increased\_activity\_energy 0.5 \\ 
        euphoria\_irritability\_expansiveness 0.5 \\ 
        racing\_thoughts 0.5 \\ 
        increased\_talkativeness 0.5 \\ 
        increased\_self\_esteem 0.5 \\ 
        diminished\_interest\_pleasure 2.0 \\ 
        reduced\_concentration 2.0 \\ 
        disrupted\_excessive\_sleep 2.0 \\ 
        change\_in\_appetite\_weight 2.0 \\ 
        psychomotor\_disturbances 2.0
        }
        & mixed 2
        & \makecell[l]{depressive \\ hypomanic}
        \\

{No. 4} & BPD2
        & \makecell[l]{
        depressed\_mood 5.7 \\ 
        diminished\_interest\_pleasure 5.7 \\ 
        reduced\_concentration 3.5 \\ 
        low\_self\_worth 2.0 \\ 
        psychomotor\_disturbances 5.7
        }
        & hypomanic 1
        & depressive
        \\

{No. 5} & BPD1
        & \makecell[l]{
        depressed\_mood 7.5 \\ 
        low\_self\_worth 7.5 \\ 
        disrupted\_excessive\_sleep 4.0 \\ 
        reduced\_energy 5.5 \\ 
        change\_in\_appetite\_weight 5.5 \\ 
        euphoria\_irritability\_expansiveness 3.0 \\ 
        increased\_activity\_energy 2.5 \\ 
        racing\_thoughts 2.5 \\ 
        decreased\_need\_for\_sleep 1.0 \\ 
        distractibility 1.0
        }
        & -
        & mixed
        \\

{No. 6} & BPD2
        & \makecell[l]{
        depressed\_mood 2.0 \\ 
        diminished\_interest\_pleasure 1.5 \\ 
        reduced\_energy 1.0
        }
        & \makecell[l]{depressive 1 \\ hypomanic 1}
        & -
        \\

{No. 7} & BPD1      
        & \makecell[l]{
        depressed\_mood 1.8 \\ 
        increased\_activity\_energy 0.5 \\ 
        reduced\_concentration 1.0
        }
        & \makecell[l]{depressive 1 \\ mixed 1}
        & -
        \\

{No. 8} & SEDD
        & \makecell[l]{
        depressed\_mood 4.0
        }
        & depressive 1
        & -
        \\

{No. 9} & SEDD      
        & \makecell[l]{
        depressed\_mood 2.5 \\ 
        recurrent\_thoughts\_death\_suicide 1.7 \\ 
        change\_in\_appetite\_weight 1.0
        }
        & depressive 1
        & -
        \\

{No. 10} & -
        & \makecell[l]{
        depressed\_mood 1.5 \\ 
        reduced\_energy 0.9 \\ 
        increased\_self\_esteem 0.7
        }        
        & -
        & -
        \\
        
{No. 11} & -        
        & \makecell[l]{
        increased\_talkativeness 1.2 \\ 
        euphoria\_irritability\_expansiveness 1.0
        }
        & hypomanic 1
        & -
        \\

{No. 12} & BPD1        
        & \makecell[l]{
        euphoria\_irritability\_expansiveness 2.5 \\ 
        increased\_activity\_energy 3.2 \\ 
        increased\_talkativeness 1.8 \\ 
        racing\_thoughts 2.9 \\ 
        decreased\_need\_for\_sleep 2.7
        }
        & hypomanic 2
        & manic
        \\

{No. 13} & BPD1     
        & \makecell[l]{
        euphoria\_irritability\_expansiveness 1.5 \\ 
        increased\_self\_esteem 1.2 \\ 
        distractibility 1.8 \\ 
        impulsive\_reckless\_behavior 2.0 \\ 
        increased\_sexual\_sociability\_goal\_directed\_activity 2.3
        }
        & manic 1
        & -
        \\

{No. 14} & BPD1     
        & \makecell[l]{
        depressed\_mood 3.5 \\ 
        diminished\_interest\_pleasure 3.1 \\ 
        euphoria\_irritability\_expansiveness 2.6 \\ 
        increased\_activity\_energy 2.4
        }
        & mixed 2
        & -
        \\

{No. 15} & BPD2     
        & \makecell[l]{
        euphoria\_irritability\_expansiveness 0.7 \\ 
        increased\_activity\_energy 1.2 \\ 
        increased\_talkativeness 1.8 \\ 
        racing\_thoughts 0.7 \\ 
        decreased\_need\_for\_sleep 1.2
        }
        & depressive 2
        & hypomanic
        \\

{No. 16} & BPD2   
        & \makecell[l]{
        depressed\_mood 2.0 \\ 
        reduced\_concentration 3.1 \\ 
        low\_self\_worth 2.7 \\ 
        increased\_activity\_energy 1.2
        }
        & \makecell[l]{depressive 1 \\ hypomanic 1}
        & -
        \\

{No. 17} & BPD1     
        & \makecell[l]{
        euphoria\_irritability\_expansiveness 2.8 \\ 
        increased\_activity\_energy 2.8 \\ 
        racing\_thoughts 3.0 \\ 
        decreased\_need\_for\_sleep 2.5 \\ 
        impulsive\_reckless\_behavior 2.7
        }
        & manic 1
        & manic
        \\

{No. 18} & RDD     
        & \makecell[l]{
        psychomotor\_disturbances 3.0 \\ 
        hopelessness 2.9 \\ 
        recurrent\_thoughts\_death\_suicide 4.0
        }
        & depressive 2
        & -
        \\

{No. 19} & BPD2      
        & \makecell[l]{
        diminished\_interest\_pleasure 3.4 \\ 
        increased\_self\_esteem 2.2 \\ 
        decreased\_need\_for\_sleep 2.4
        }
        & \makecell[l]{hypomanic 1 \\ depressive 1}
        & -
        \\

{No. 20} & SEDD
        & \makecell[l]{
        delusions 4.1 \\ 
        passivity\_experiences 3.7 \\ 
        disorganized\_behavior 3.9
        }
        & depressive 1
        & -
        \\

{No. 21} & BPD1     
        & \makecell[l]{
        reduced\_energy 2.5 \\ 
        disrupted\_excessive\_sleep 3.0 \\ 
        change\_in\_appetite\_weight 2.8 \\ 
        psychomotor\_disturbances 2.9
        }
        & \makecell[l]{manic 1 \\ depressive 1}
        & -
        \\

{No. 22} & BPD1     
        & \makecell[l]{
        depressed\_mood 3.6 \\ 
        hopelessness 2.8 \\ 
        increased\_activity\_energy 3.2 \\ 
        impulsive\_reckless\_behavior 3.1
        }
        & \makecell[l]{mixed 1 \\ hypomanic 1}
        & -
        \\

{No. 23} & BPD2     
        & \makecell[l]{
        euphoria\_irritability\_expansiveness 0.5 \\ 
        increased\_activity\_energy 0.5 \\ 
        increased\_self\_esteem 0.5 \\ 
        impulsive\_reckless\_behavior 0.5 \\ 
        distractibility 0.5
        }
        & depressive 1
        & hypomanic
        \\

{No. 24} & -    
        & \makecell[l]{
        increased\_activity\_energy 2.6 \\ 
        distractibility 2.3 \\ 
        racing\_thoughts 2.7 \\ 
        increased\_self\_esteem 2.9 \\ 
        impulsive\_reckless\_behavior 2.6
        }
        & -
        & -
        \\

{No. 25} & RDD   
        & \makecell[l]{
        low\_self\_worth 2.3 \\ 
        recurrent\_thoughts\_death\_suicide 3.8 \\ 
        change\_in\_appetite\_weight 2.7
        }
        & depressive 2
        & -
        \\

{No. 26} & BPD2      
        & \makecell[l]{
        depressed\_mood 5.7 \\ 
        diminished\_interest\_pleasure 5.7 \\ 
        reduced\_concentration 3.5 \\ 
        low\_self\_worth 2.0 \\ 
        psychomotor\_disturbances 5.7 \\ 
        euphoria\_irritability\_expansiveness 0.5 \\ 
        increased\_activity\_energy 0.5 \\ 
        increased\_self\_esteem 0.5 \\ 
        impulsive\_reckless\_behavior 0.5 \\ 
        distractibility 0.5
        }
        & \makecell[l]{depressive 1 \\ hypomanic 1}
        & \makecell[l]{depressive \\ hypomanic}
        \\

{No. 27} & SEDD  
        & \makecell[l]{
        reduced\_concentration 3.5 \\ 
        low\_self\_worth 2.0 \\ 
        hopelessness 5.7 \\ 
        recurrent\_thoughts\_death\_suicide 4.0 \\ 
        disrupted\_excessive\_sleep 3.5 \\ 
        change\_in\_appetite\_weight 2.0 \\ 
        psychomotor\_disturbances 5.7 \\ 
        reduced\_energy 4.0
        }
        & depressive 1
        & -
        \\

{No. 28} & SEDD
        & \makecell[l]{
        depressed\_mood 5.7 \\ 
        diminished\_interest\_pleasure 5.7 \\ 
        reduced\_concentration 3.5 \\ 
        low\_self\_worth 2.0 \\ 
        psychomotor\_disturbances 5.7
        }
        & -
        & depressive
        \\

{No. 29} & RDD     
        & \makecell[l]{
        depressed\_mood 5.7 \\ 
        diminished\_interest\_pleasure 5.7 \\ 
        reduced\_concentration 3.5 \\ 
        low\_self\_worth 2.0 \\ 
        psychomotor\_disturbances 5.7
        }
        & depressive 1
        & depressive
        \\

{No. 30} & -        
        & \makecell[l]{
        euphoria\_irritability\_expansiveness 0.7 \\ 
        increased\_activity\_energy 1.2 \\ 
        increased\_talkativeness 1.8 \\ 
        racing\_thoughts 0.7 \\ 
        decreased\_need\_for\_sleep 1.2
        }
        & -
        & hypomanic
        \\

\end{longtblr}

\clearpage

\section{Prompts}
\label{sec:appendix_prompts}

This section shows the full prompts we used to interact with the LLMs.
Whenever applicable, we used system and user prompts as follows.

\subsection{Translating ICD-11 CDDR Manual to Datalog Rules}

\begin{quote}
\emph{\footnotesize
\textbf{System}:
You are an expert at translating mental health diagnostic criteria into Soufflé Datalog code.
Translate the given criterion into a .dl program using Soufflé syntax as follows.
The patient information is given as input to the program as \texttt{Observed} and \texttt{History} relations.
The patient diagnosis is returned as output from the program as \texttt{Diagnosis} relation.
\begin{itemize}
\item \texttt{.decl Observed(Patient:symbol, Symptom:symbol, Week:float)} describes that Patient has experienced Symptom for Week number of weeks.
\item \texttt{.decl History(Patient:symbol, Condition:symbol, Count:number)} describes that Patient has experienced Condition for Count number of times.
\item \texttt{.decl Diagnosis(Patient:symbol, Disorder:symbol)} describes that Patient has been diagnosed with Disorder.
\end{itemize}
}

\emph{\footnotesize
For context, here is an example of Scizophrenia criterion translated into Soufflé .dl code.
\begin{itemize}
\item Scizhophrenia criterion: \textcolor{darkblue}{[Scizhophrenia criterion from ICD-11 CDDR]}.
\item Relevant symptom names for \texttt{Observed} relation:
\textcolor{darkblue}{[Symptom names]}
\item Soufflé .dl code: \textcolor{darkblue}{[Manually crafted Datalog program for Schizophrenia]}
\end{itemize}
}

\emph{\footnotesize
\textbf{User}:
Now, translate the following criteria into Souffle .dl code for Bipolar I, Bipolar II, Single Episode Depressive Disorder, and Recurrent Depressive Disorder.
\begin{itemize}
\item Mood Episode criterion: \textcolor{darkblue}{[Depressive, Manic, Mixed, and Hypomanic Episode criteria from ICD-11 CDDR]}.
\item Mood Disorder criterion: \textcolor{darkblue}{[Bipolar I, Bipolar II, Single Episode Depressive Disorder, and Recurrent Depressive Disorder criteria from ICD-11 CDDR]}.
\item Relevant symptom names for \texttt{Observed} relation:
\textcolor{darkblue}{[Symptom names]}
\item Relevant condition names for \texttt{History} relation:
\textcolor{darkblue}{[Condition names]}
\end{itemize}
}

\end{quote}

\subsection{Generating Diagnosis by \emph{LLM-only Approach}}

\begin{quote}
\emph{\footnotesize
\textbf{System}: 
You are an expert at diagnosing patients according to the ICD-11 Clinical Descriptions and Diagnostic Requirements (CDDR).
The patient data are represented by a list of current symptoms denoted as \texttt{Observed} and a list of history denoted as \texttt{History}.
\texttt{Observed} matches the patient with the symptom and the number of weeks it has been observed.
\texttt{History} matches the patient with the condition and the number of times it existed.
No record for a patient means that there is no related data for them.
The considered disorders are: Bipolar I, Bipolar II, Single Episode Depressive Disorder, and Recurrent Depressive Disorder.
}

\emph{\footnotesize
\textbf{User}:
For brevity, please output only the diagnosis for the following patients.
Patients with no clear diagnosis should be indicated as such.
\begin{itemize}
\item \texttt{Observed}: \textcolor{darkblue}{[Observed Data]}
\item \texttt{History}: \textcolor{darkblue}{[History Data]}
\end{itemize}
}
\end{quote}